\documentclass[runningheads]{llncs}
\usepackage{graphicx}
\usepackage{hyperref}

\usepackage{amsfonts}
\usepackage{mathtools}
\usepackage{tikz}
\usetikzlibrary{matrix}
\usepackage{float}
\usepackage{subcaption}
\usepackage[misc,geometry]{ifsym}

\begin{document}
\title{Synthesising Reinforcement Learning Policies through Set-Valued Inductive Rule Learning}
\titlerunning{Synthesising RL Policies through Set-Valued Inductive Rule Learning}
\author{Youri Coppens\inst{1,2}\orcidID{0000-0003-1124-0731} \Letter \and \\
Denis Steckelmacher\inst{1}\orcidID{0000-0003-1521-8494} \and \\
Catholijn M. Jonker\inst{3,4}\orcidID{0000-0003-4780-7461} \and \\
Ann Nowé\inst{1}\orcidID{0000-0001-6346-4564}}
\authorrunning{Y. Coppens et al.}
%
\institute{Vrije Universiteit Brussel, Brussels, Belgium \and
Université Libre de Bruxelles, Brussels, Belgium \and
Delft University of Technology, Delft, The Netherlands \and
Leiden Institute of Advanced Computer Science (LIACS), Leiden, The Netherlands\\
\email{yocoppen@ai.vub.ac.be}}
\maketitle    
\begin{abstract}
Today's advanced Reinforcement Learning algorithms produce black-box policies, that are often difficult to interpret and trust for a person.
We introduce a policy distilling algorithm, building on the CN2 rule mining algorithm, that distills the  policy into a rule-based decision system.
At the core of our approach is the fact that an RL process does not just learn a policy, a mapping from states to actions, but also produces extra meta-information, such as action values indicating the quality of alternative actions. This meta-information can indicate whether more than one action is near-optimal for a certain state.
We extend CN2 to make it able to leverage knowledge about equally-good actions to distill the policy into fewer rules, increasing its interpretability by a person. Then, to ensure that the rules explain a valid, non-degenerate policy, we introduce a refinement algorithm that fine-tunes the rules to obtain good performance when executed in the environment.
We demonstrate the applicability of our algorithm on the Mario AI benchmark, a complex task that requires modern reinforcement learning algorithms including neural networks. The explanations we produce capture the learned policy in only a few rules, that allow a person to understand what the black-box agent learned. Source code: \url{https://gitlab.ai.vub.ac.be/yocoppen/svcn2}.

\end{abstract}
\section{Introduction}\label{sec:intro}
Reinforcement Learning (RL) is a machine learning technique that learns from experience, rather than data that has been collected offline. 
By interacting with its environment, an intelligent agent learns to adapt its actions to show the desired behaviour.  
This environment is in general stochastic and its dynamics unknown to the agent.

RL is particularly useful in settings without a model of the environment, or when the environment is hard to model. 
RL is also interesting when an accurate simulation model can be built, but solving that model to obtain a decision or control strategy is hard.
Telecommunication applications such as routing, load balancing and call admission control are examples where RL has proven its worth as a machine learning technique.
RL has unfortunately one crucial weakness which may prevent it from being applied in future critical applications: the control policies learned by the agent, especially using modern algorithms including neural networks, are black boxes that are hard to interpret.
This shortcoming could prove a major obstacle to the adoption of RL.
For example, the new EU privacy regulation, more specifically Article 22 of the General Data Protection Regulation (GDPR), requires that all AI with an impact on human lives needs to be accountable, which, for an RL agent, means that it must be possible to tell the user why the agent took a particular action in a state.
Interpretability of machine learning based models is therefore crucial, as it forms the basis for other concerns such as accountability, that subsumes the justification of the decisions taken.
Assume that RL is used to synthesise a prevention strategy for a dangerous infection disease. 
The agent should learn what regulations and procedures people should follow, while optimally allocating a budget, such as vaccines, medication or closed-school days.
While this scenario is plausible, see~\cite{Libin2020SEIR}, will a black-box strategy be accepted?  
The above observation brings us to the following key question:
Wouldn't it make more sense to use RL to explore possible strategies and simulate the potential effects on our society, before we would decide to implement such a strategy? However, this requires that the strategies found by RL algorithms are presented in a human-understandable format.

Explainable AI, reviewed in~\cite{miller_explanation_2019}, is not new.
Even in the early days of Expert Systems, many approaches have been proposed to allow to automatically generate explanations and to reflect on the reasoning being used, see e.g.~\cite{maes_computational_1987}. 
Recently, explainability has been put more prominently on the agenda in the Machine Learning (ML) community as well~\cite{molnar_interpretable_2019}.
The massive success of the black-box ML-techniques has culminated in many applications that display powerful results, but provide poor insight into how they find their conclusion. 
Over the last three years, workshops at major conferences in Machine Learning (e.g.~ICML, NeurIPS), Artificial Intelligence (e.g.~IJCAI), Human-Computer Interaction (e.g.~ACM CHI, ACM IUI) and Planning (e.g.~ICAPS) have been held for contributions that increase human interpretability of intelligent systems.
At these workshops, initial research related to data mining has been proposed. 
However, ideas to make RL more explainable are rather limited and often borrowed from existing ML Interpretability literature
\cite{alharin_reinforcement_2020}.

In this paper, we present a method that allows the policy learned by a reinforcement learning agent, using modern deep learning methods, to be expressed in a way that is understandable by a person. Our contributions are as follows:

\begin{enumerate}
    \item We extend the CN2 rule-mining algorithm to make it able to leverage the meta-information of an RL process on equally-good actions. This allows the algorithm to map an RL policy to a simpler and compacter set of rules.
    \item We use our extended CN2 algorithm to produce human-understandable rules from a black-box RL policy, and evaluate our method on the Mario benchmark.
    \item We introduce a 2-phased method, that produces a first set of rules providing a global view on the policy. Then refines the rules when needed. This refinement is driven by the performance of the rules when used to perform the task, rather than the accuracy when mining the rules.
\end{enumerate}

\section{Related Work}\label{sec:related-work}
Until now, insights in the learnt RL policy were typically limited to showing a typical trace, i.e.~an execution of the policy or through analysing and visualising agent reward structures~\cite{agogino_analyzing_2008}. 
Relevant efforts to mention are~\cite{rucksties_exploring_2010}, where the authors visualise action versus parameter exploration and~\cite{Brys2014} that analyses and visualises the usefulness of heuristic information in a state space.

Distillation techniques translate the behaviour from large complex models to simpler surrogate models~\cite{Hinton2015distill}.
While these methods reduce computational overhead and model complexity, the general focus is mostly set on maintaining performance rather than increasing the interpretability through the surrogate model, even in the context of RL policies~\cite{Rusu2016}.
In~\cite{Coppens2019a}, a deep RL policy is distilled into a surrogate soft decision tree model~\cite{Frosst2017}, in an effort to gain insights in the RL policy.
The surrogate model determines a hierarchy of filters that perform sequential feature extraction from observations.
These filters highlight which features are considered important in the decision making.
Our work is complementary in the sense that the rules we extract look at the content, i.e.~the actual value of the observation features, rather than the importance of a particular feature.

Relational RL (RRL) could be seen as an interpretable RL approach \emph{by design}~\cite{tadepalli_relational_2004} as it allows to express the policy as well as the quality of the policy using relational expressions.
Relational RL requires domain-specific knowledge from the designer, as a set of concepts have to be defined \textit{a priori}. If these concepts are not carefully defined, the expressiveness of the agent is reduced, which may prevent it from learning any good policy. 
In~\cite{zambaldi2018deep}, relational inductive biases have been incorporated into a deep learning process, in an attempt to scale RRL without pre-defined concepts, thus allowing for relevant concepts to emerge. 
The approach is shown to reveal the high-level plan of the RL strategy, thus identifying subgoals and options that can be generalised.
Our work does not consider subgoals and instead focuses on making the actual policy, the mapping from states to actions, transparent to the user.

The paper by Madumal et al.~\cite{madumal_explainable_2020} introduces an approach to explain a policy which can either be obtained through model-based RL or planning. 
It builds upon causal graphs as a means to calculate counterfactuals. 
As the causal graph is assumed to be given, the symbolic concepts are also defined \emph{a priori}, and have to be carefully chosen, leading to the same vulnerability as Relational RL.

In our approach, we don't want to restrict the policy \emph{a priori}.
We allow the RL process to learn complex policies if needed. 
Recent advances in (deep) function approximation for RL have yielded powerful techniques that learn rich state representations and have advanced state-of-the-art in terms of learning performance in complex environments, see e.g.~\cite{Mnih2015}. 
However, these techniques obscure a proper understanding of the policy that is being learned.

In the approach we propose, we start from existing state-of-the-art RL techniques and make them more interpretable. 
Our work performs an \emph{a posteriori} simplification and communication of the strategy. 
This is realised through a rule-mining approach that exploits the meta-information that is present in the RL learning algorithm. 
While decision trees are a more commonly adopted interpretable surrogate model that can be visualised in a natural manner, we opted for decision rules, because these have a similar expressiveness while also being more compact and robust.
In addition, the use of meta-information, such as which actions are equally good in a state, is more naturally implemented in a rule-mining system, as well as the refinement process we propose.
The RL process not only provides us with the best action for a given state, but the probability that an RL actor gives to each of the actions is providing us information on the quality of the sub-optimal actions.
Exploiting this meta-information allows to go beyond a mere translation of the policy into a set of rules. 
As, for instance, knowing that two actions have nearly the same probability in a given state indicates that they are both \textit{equally good}, which allows our rule mining algorithm to chose one action or the other, depending on what leads to the simplest rule set.
This exploitation of the meta-information of an RL process also differentiates our work from approaches that translate the RL policy in a set of fuzzy rules, such as~\cite{huang_interpretable_2020}~and~\cite{gevaert_distillation_2019}.

\section{Background}\label{sec:background}
Our work covers several domains of research such as Reinforcement Learning and Inductive Rule Learning.

    \subsection{Policies learnt through Reinforcement Learning}\label{subsec:rl}
    Reinforcement Learning (RL) tackles sequential decision making problems in an interactive setting.~\cite{Sutton2018}
    An artificial agent learns by operating in an unknown environment, from scratch or boosted by some initial domain knowledge, which actions are optimal over time. 
    Based on the observation of the state of the environment, the agent executes an action of choice which makes the agent transition to a next state.
    In addition to the new state, the agent will also receive feedback from the environment in the form of a numerical reward.
    Based on that feedback, the agent will adapt its behaviour into a desired strategy.
    
    Reinforcement learning problems are formally represented as a Markov Decision Process (MDP), a 5-tuple $(S,A,T,R,\gamma)$, with $S$ and $A$ the sets of states and actions available in the environment.
    Unknown to the agent are $T$ and $R$.
    $T: S \times A \times S \rightarrow [0,1]$ represents the transition probability function, assigning probabilities to a possible transition from state $s \in S$ to another state $s' \in S$, by executing action $a \in A$.
    $R: S \times A \rightarrow \mathbb{R}$ is the environment's reward function, providing the agent with feedback and finally a discount factor $\gamma \in [0, 1]$ regulates the agent's preference for immediate rewards over long-term rewards.
    The agent's goal is to maximise the expected return $\mathbb{E}[G_t]$, where $G_t=\sum_{i=0}^{\infty} \gamma^iR_{t+i}$, the sum of future discounted rewards.
    Lower values of $\gamma$ set the preference on immediate reward, whilst higher values result in a more balanced weighing of current and future reward in $G_t$.
    The agent learns a policy $\pi: S \times A \rightarrow [0,1]$, mapping state-action pairs to probabilities, describing the optimal behaviour strategy.

    In value-based RL methods, this policy is implicitly present through the use of a \emph{value function} in combination with an action-selection strategy.
    The value function $Q(s,a)$ is often represented as the expected return when taking an action $a$ in state $s$: $Q(s,a) = \mathbb{E}[G_t | S_t = s, A_t = a]$, hence describing which actions are expected to result in more future rewards.

    When the Q-values have converged, the highest ranked action is the optimal one, whilst others are granted to be sub-optimal. However, when the Q-values are equal or almost equal, this indicates that there is no clear preference over these actions.

    Policy-based algorithms are another family of RL methods. They maintain an explicit policy, or \textit{actor}, often described as a parametric expression $\pi_\theta$. 
    The policy is optimised iteratively, using methods such as computing and following the gradient of its quality~\cite{Sutton2000}, or training it to progressively imitate the greedy policy of a collection of critics~\cite{Steckelmacher2019}. Policy-based methods typically learn a stochastic policy where the difference in probability can be considered to be a proxy for the difference in quality of two alternative actions in a state. The BDPI algorithm~\cite{Steckelmacher2019} that we use in this paper, for instance, learns a policy that associates nearly-equal probability to actions that are of comparable quality, while actions that are clearly worse than the other ones have a probability near 0. In other words, when the probability of two actions in a given state is close to each other, this expresses there is no clear preference of one action over the other action.
    We exploit this property in our framework in order to find a small and simple set of rules capturing the main elements of the policy.

    \subsection{Inductive Rule Learning}\label{subsec:rule-learning}
    Inductive rule learning allows us to extract knowledge in a symbolic representation from a set of instances in a supervised manner.
    The goal of supervised rule induction is the discovery of rules that reflect relevant dependencies between classes and attribute values describing the given data~\cite{furnkranz_foundations_2012}. 
    In our setting, these data instances are state-action pairs drawn from a learned RL policy.
    More precisely, the instances are pairs of states with \emph{set-valued} actions:
    for a single instance, a state might be paired with multiple actions.
    
    The set-valued aspect is important to express that two or more actions are high-quality alternatives in a certain state. 
    By supporting the possibility to express this indifference towards which action of that set is selected in the corresponding state, we can create rules with a larger support and as a consequence fewer rules.
    Regular inductive rule learning is not designed to deal with this set-valued aspect, and therefore we extend this learning paradigm in Section \ref{sec:set-valued-rule-mining}.
    Before we introduce our set-valued rule mining, we first summarise the main elements of inductive rule learning and the CN2 algorithm.

    Inductive rule learning allows to summarise a set of instances, which might be discrete valued or continuous valued, along with a discrete target class, into a set of rules. 
    The rules are of the form `IF $Conditions$ THEN $c$', where $c$ is the class label. 
    Because of their symbolic nature, the resulting model is easy to interpret by humans.

    We base our work on CN2~\cite{clark_cn2_1989}, a frequently used rule learning algorithm based on the separate-and-conquer principle, and extend it to support multiple equally-good labels per input. 
    In CN2, rules are iteratively identified through the application of beam search, a heuristic tree search algorithm.
    The best rule that is identified in this search procedure will be selected as new rule.
    The consequent (classifying label) of the rule is the majority class of the examples covered by the rule.

    The commonly used evaluation heuristic in CN2 is \emph{weighted relative accuracy} (WRA)~\cite{lavrac_rule_1999,todorovski_predictive_2000}, which aims to find a good balance between a large coverage and a high accuracy.
    Equation~\ref{eq:wra} describes the WRA of a proposed rule $r$: 
    \begin{equation} \label{eq:wra}
        \text{WRA}(r) = \frac{\hat{P}(r) + \hat{N}(r)}{P(r)+N(r)} \times \left(\frac{\hat{P}(r)}{\hat{P}(r)+\hat{N}(r)} - \frac{P(r)}{P(r)+N(r)} \right)
    \end{equation}
    where $P(r)$ and $N(r)$ represent the number of positive and negative examples in the current training set and $\hat{P}(r)$ and $\hat{N}(r)$ represent the number of positive and negative examples covered by the considered rule $r$.
    From hereon, we will drop the parameter $r$ as no confusion is to be expected.
    Examples are said to be positive when they have the same classification label as the consequent of the rule, otherwise they are considered to be negative. 
    $\hat{P}$ and $\hat{N}$ can be thus considered the true and false positive examples covered by the rule respectively.
    The first factor in Equation~\ref{eq:wra}, $\frac{\hat{P} + \hat{N}}{P+N}$, represents the \emph{coverage} of the proposed rule, the rate of examples that satisfy the antecedent, and weighs the second factor of the equation.
    That factor, $\left(\frac{\hat{P}}{\hat{P}+\hat{N}} - \frac{P}{P+N} \right)$, represents the relative classification accuracy of a rule. 
    It is the difference between the accuracy of the samples covered by the rule and the overall accuracy of the training examples.

    The algorithm stops generating rules if there are no other significant rules to be found or an insufficient number of samples remain.
    These thresholds are expected to be application dependent and are therefore to be determined by the user.
    
    We now introduce the main contribution of this paper, a rule mining algorithm based on CN2 that extends two aspects of it relevant for reinforcement learning applications: allowing states to be mapped to \emph{several} equally-good actions, and ensuring that the distilled policy performs comparably to the original policy.

\section{Extending CN2 to Set-Valued Labels}\label{sec:set-valued-rule-mining}
Our main contribution consists of extending two important aspects of CN2, to make it suitable to policy distillation in an explainable reinforcement learning setting:

\begin{enumerate}
    \item We exploit meta-information generated by the reinforcement learning process to associate \textit{sets} of equally-good actions to each state. By offering choice to the rule miner in which action to map to every state, we potentially (and in practice do) reduce the number of rules needed to cover the original policy.
    \item Contrary to classical Supervised Learning settings, the main application domain of CN2, what matters in our setting is not primarily the classification accuracy, but the loss in performance of the distilled policy. In other words, one little misclassification in a particular state can have a big impact on the performance of the policy, but in other states, this misclassification might almost be unnoticeable from a policy performance perspective. We propose a two-step approach that allows to identify those \textit{important} misclassifications, leading to short rule sets that, when executed in the environment, still perform as well as the original policy. This ensures that the explanations we provide with the mined rules correspond to a good policy, and not some over-simplified degenerate policy.
\end{enumerate}

The first point, associating sets of equally-good actions to each state, requires an extension of the CN2 algorithm itself. The second point, and the application of the whole framework to reinforcement learning policy distillation, requires an iterative framework that we present in Section \ref{sec:methods}.
    
The CN2 algorithm iteratively learns a rule through a beam search procedure. 
The objective of this step is to search for a rule that on the one hand covers a large number of instances, and on the other hand has a high accuracy.
This is steered through the evaluation heuristic, i.e.~the weighted relative accuracy (WRA), which makes a distinction between positive and negative examples based on the label of the training sample.
In order to deal with the Set-Valued instances, which express an indifference with respect to the actual label chosen by the rule, we adapted the WRA heuristic. 

In the regular single-label setting, the sum of positive and negative examples covered by the rule and in the overall training set, $\hat{P}+\hat{N}$ and $P+N$, always corresponds to the number of respective training examples, $\hat{E}$ and $E$.
However, in a multi-label setting this is not necessarily the case, as an instance can be assigned to any of the labels and is thus potentially both a positive and negative example at the same time according to the standard WRA heuristic.
The heuristic aims at maximising the correctly classified instances for a rule by assigning the majority class of the covered samples as the class label.
In order to deal with the indifference to which instances with multiple class labels are assigned, we propose the following adaptation to the WRA heuristic, described in Equation~\ref{eq_wra-set}, where we replace the sum of positive and negative examples covered by the rule and in the overall set by the actual number of distinct samples in the set, respectively denoted by $E$ and $\hat{E}$:
\begin{equation} \label{eq_wra-set}
   \text{WRA}_{set} = \frac{\hat{E}}{E} \times \left( \frac{\hat{P}}{\hat{E}} - \frac{P}{E} \right)
\end{equation}

This will have the effect that the heuristic will only count the multi-labelled instances as a positive example when the majority class is one of the options, and as a negative example otherwise.

\subsection{Impact of Set-Valued Labels on Policy Distillation}

\begin{figure}[t]
    \centering
    \begin{subfigure}[t]{0.49\textwidth}
         \centering
         \includegraphics[width=0.7\textwidth]{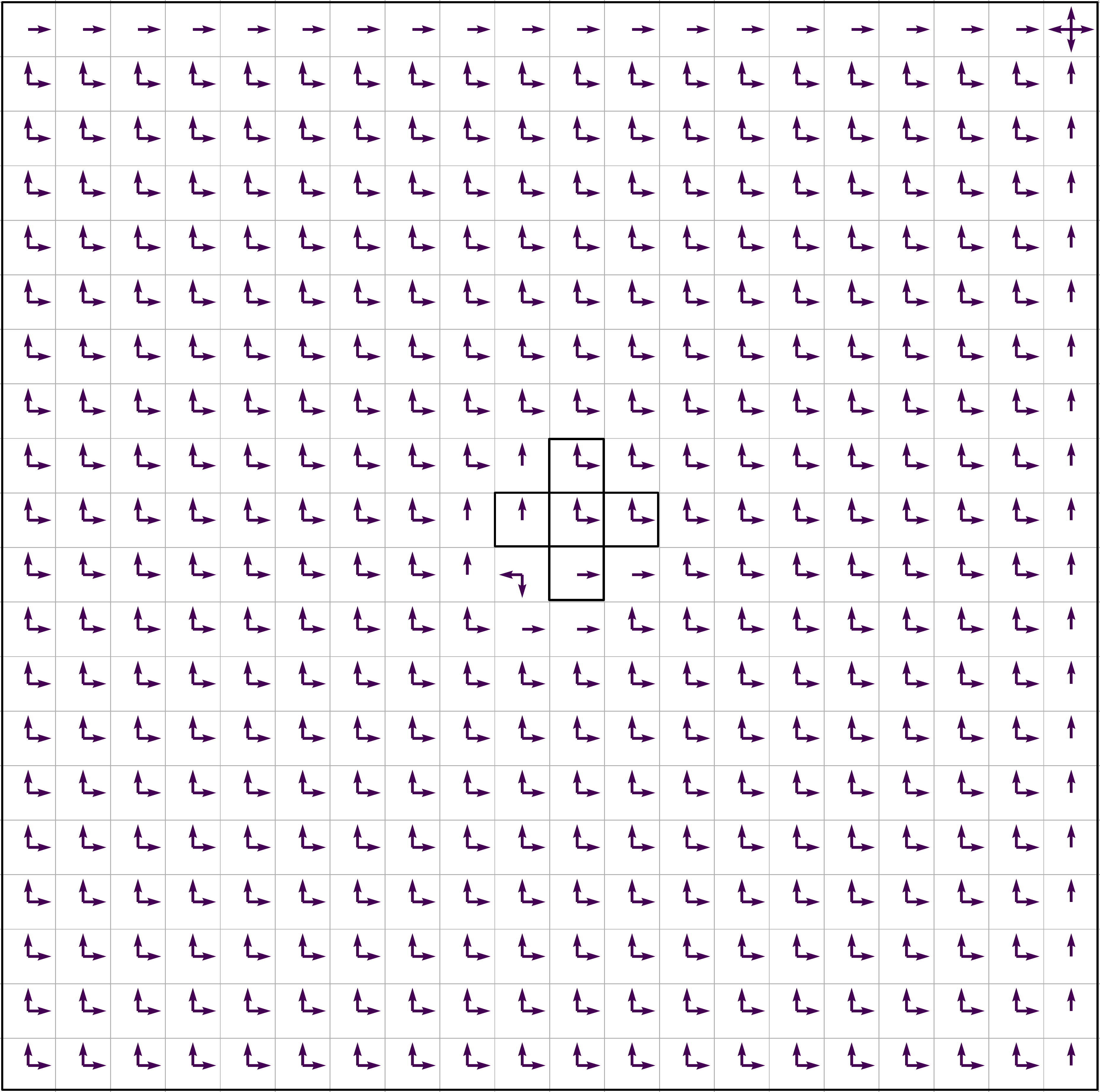}
         \caption{In each cell, the most optimal action(s) to perform are shown by arrows. Notice how some cells in the grid contain two optimal actions leading to the top rightmost goal state. The cross in the middle indicates the muddy states}
         \label{fig:grid-qlearning}
    \end{subfigure}
    \hfill
    \begin{subfigure}[t]{0.49\textwidth}
         \centering
         \includegraphics[width=0.7\textwidth]{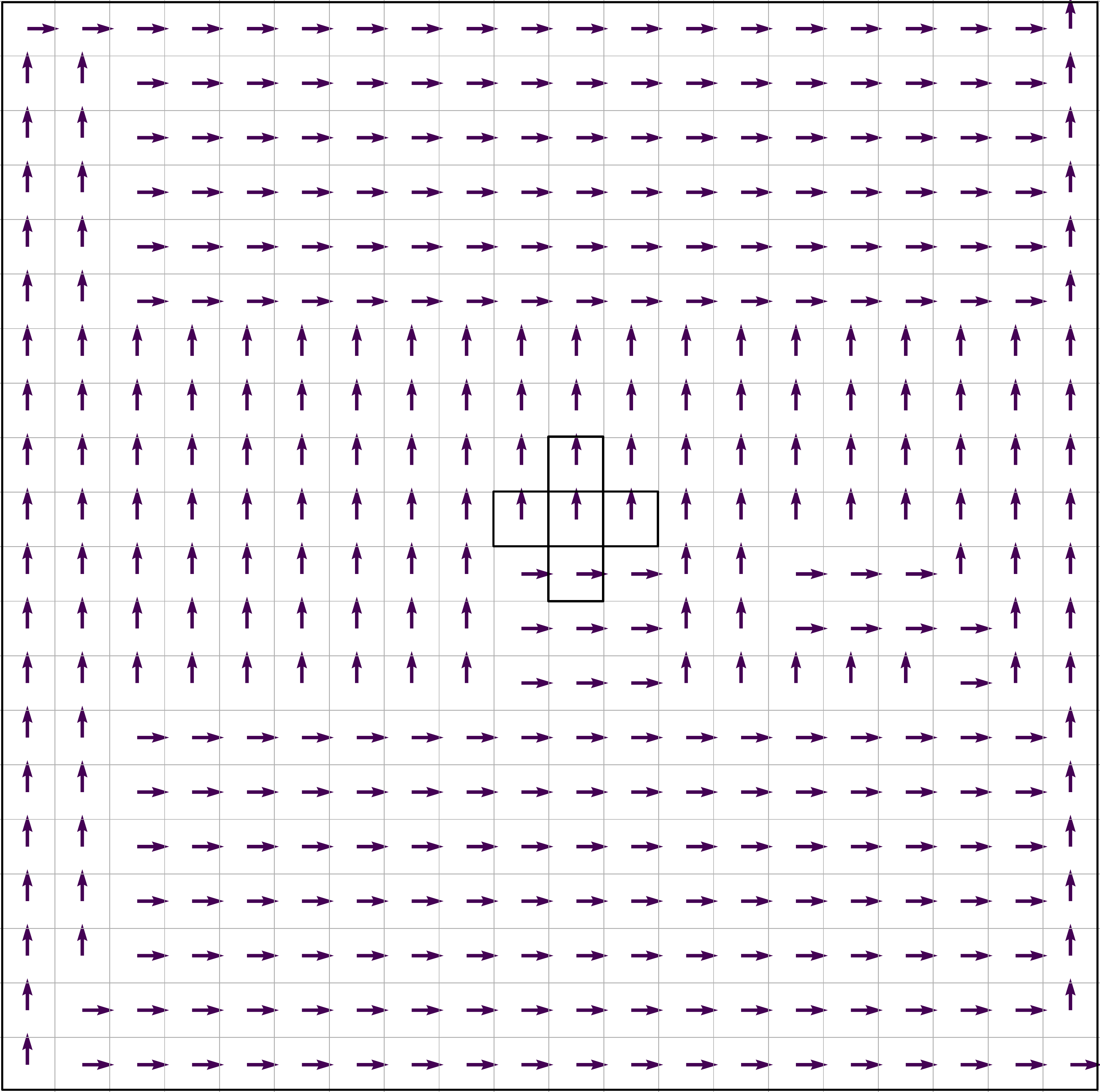}
         \caption{Distillation of the muddy gridworld policy with standard CN2: Samples provided to CN2 are state-action pairs. When multiple optimal actions exist for a state, an action is selected at random.}
         \label{fig:grid-standard-cn2}
    \end{subfigure}
    \caption{Muddy gridworld environment with its optimal policy, and a distillation of this policy using standard CN2 (no support for equally-good actions).}
\end{figure}

\begin{figure}[t]
    \centering
    \begin{subfigure}[t]{0.49\textwidth}
        \centering
        \includegraphics[width=0.7\textwidth]{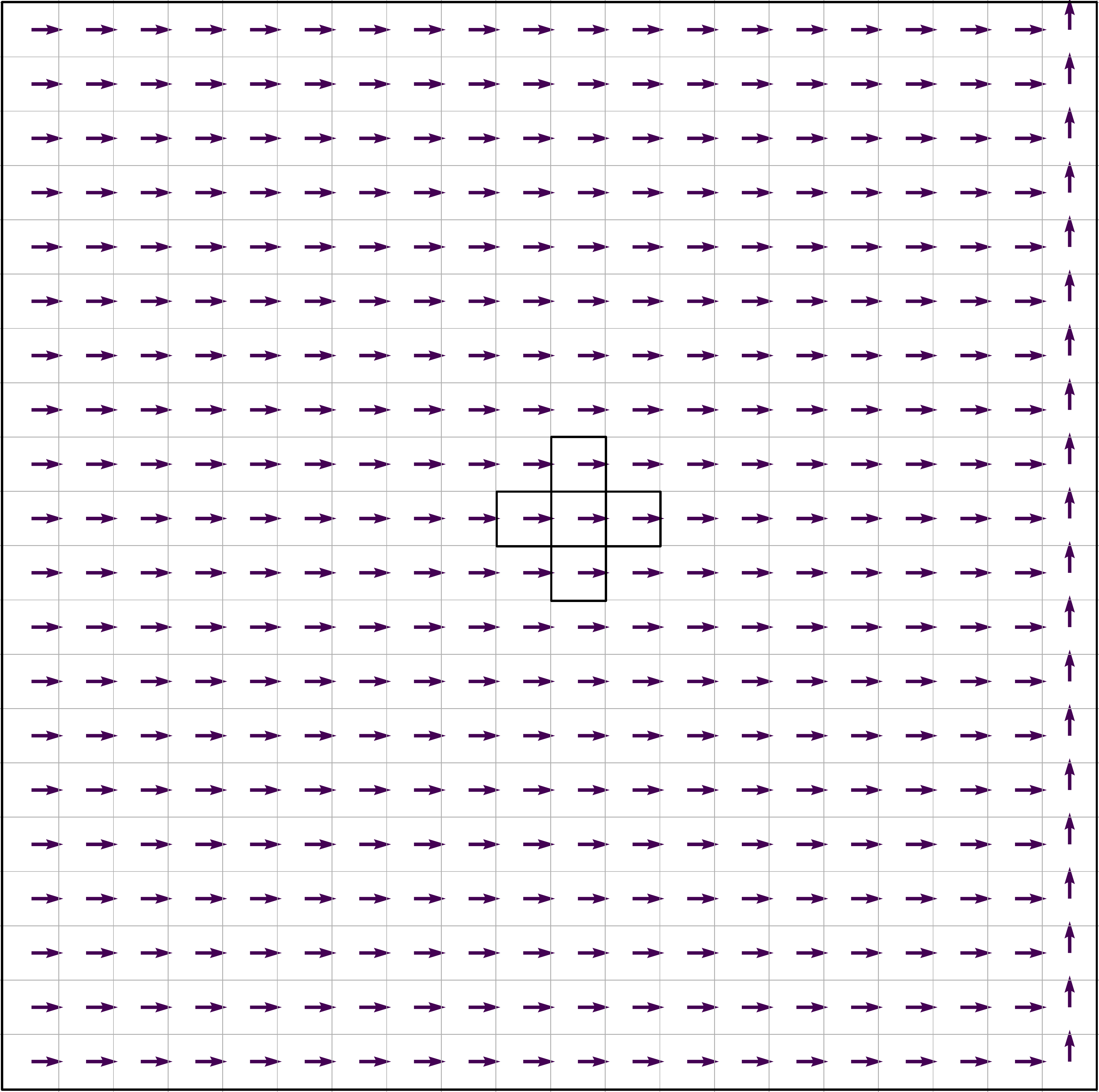}
        \texttt{
        \begin{enumerate}
            \item IF X<=18 THEN Class=RIGHT 
            \item IF X=19 THEN Class=UP
        \end{enumerate}
        }
        \caption{An initial distillation of the muddy gridworld policy using our Set-Valued extension of CN2, taking multiple possible actions into account. The rules are simple, but do not yet consider the policy details in the muddy states into account, leading to sub-optimal performance.}
        \label{fig:grid-phase1}
    \end{subfigure}
    \hfill
    \begin{subfigure}[t]{0.49\textwidth}
        \centering
        \includegraphics[width=0.7\textwidth]{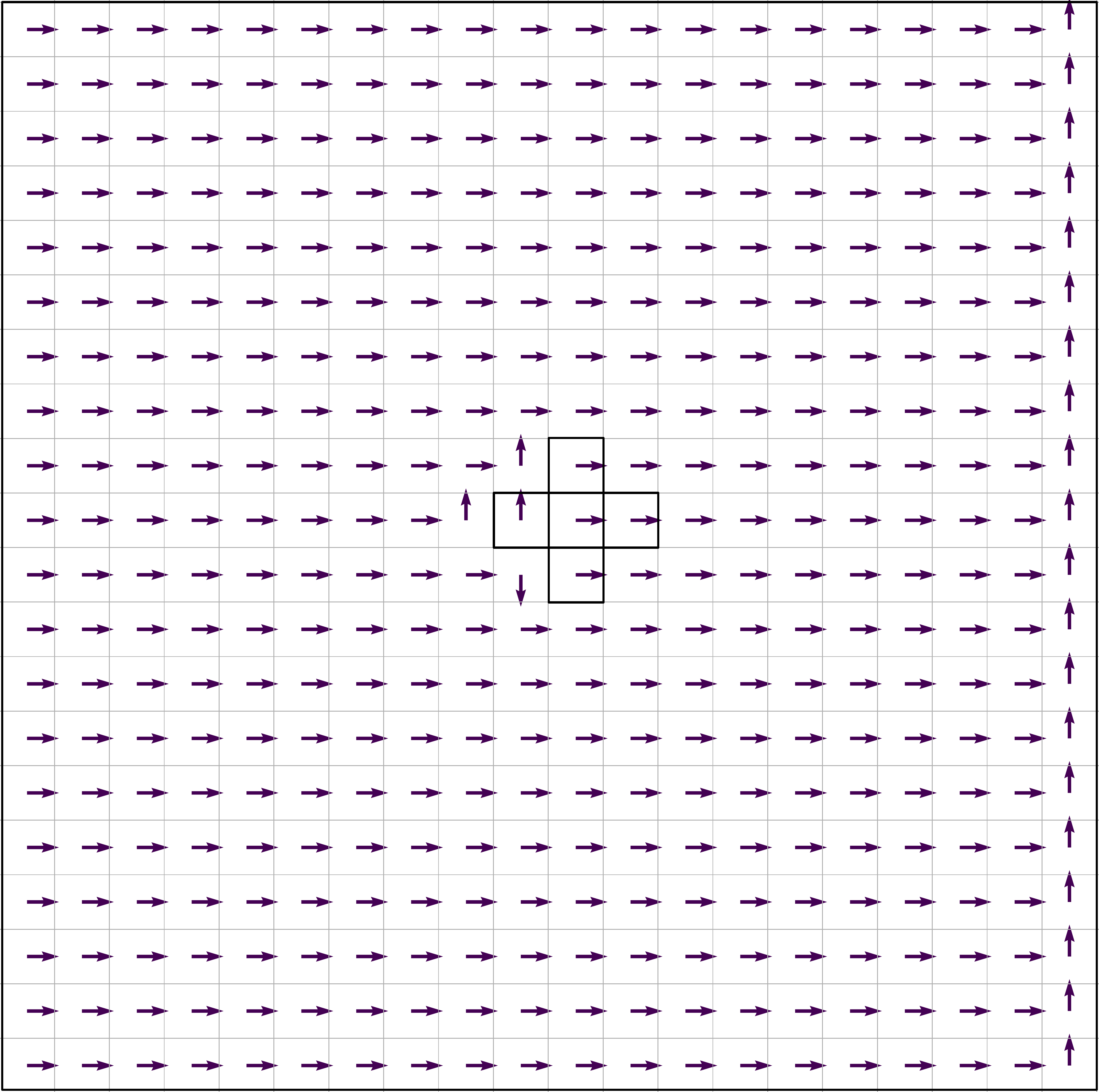}
        \texttt{
        \begin{itemize}
            \item[1.1] IF X<=18 AND Y=10 AND X>=8 AND X<=9 THEN Class=UP
            \item[1.2] IF X<=18 AND X>=10 THEN Class=RIGHT 
            \item[1.3] IF X<=18 AND X<=8 THEN Class=RIGHT 
            \item[1.4] IF X<=18 AND Y=11 THEN Class=UP 
            \item[1.5] IF X<=18 AND Y=9 THEN Class=DOWN 
            \item[1.6] IF X<=18 THEN Class=RIGHT
            \item[2] IF X=19 THEN Class=UP
        \end{itemize}
        }
         \caption{Refining the rules shown in (a), as detailed in Section \ref{sec:methods}, identifies the state near the mud that requires a different action from the surrounding states, leading to optimal behavior.}
         \label{fig:grid-phase2}
    \end{subfigure}
    \caption{Distilled policy with our modified CN2 algorithm, without and with our refinement step.}
\end{figure}

\setcounter{footnote}{0}
Before we discuss our approach on the Mario AI benchmark, we illustrate the effect of our set-valued rule mining heuristic on a simple 20 by 20 discrete-state gridworld problem, where a navigating agent has to walk to the top rightmost cell from any initial position.\footnote{Another illustration can be found on pages 8 and 9 of  \url{https://www.ida.liu.se/~frehe08/tailor2020/TAILOR_2020_paper_48.pdf}} Actions are up, down, right and left.
Near the center of the grid, a couple of cells contain mud. The agent receives a reward of -1 per time-step, -10 when entering a muddy cell.

Figure~\ref{fig:grid-qlearning} shows the environment, and the optimal policy learned by a Q-learning agent in this setting.
In most cells, two optimal actions exist to navigate to the goal, namely moving upward or to the right. Meaning there are quite a lot of optimal policies in this example.
In the cells neighboring the muddy cells however, the agent has learned not to walk towards the muddy parts, but rather walk around it.
In essence, the general behavior of the agent can be summarized as moving upward or to the right.
Hence an acceptable simplified policy would be to first walk to either the upper or right border and then walk straight to the goal.
Such a simplified policy would be a desired outcome of the rule mining process, as it provides a first high level view on the policy. This high level policy can be refined, see Section~\ref{sec:methods} below. As explained in that section, this refinement is not driven by increasing the accuracy requirements of the rule-mining algorithm, but rather by improving the performance of the distilled policy.

Figure~\ref{fig:grid-standard-cn2} shows the resulting policy derived from the standard single-labeled CN2 algorithm, using the regular WRA heuristic.
As the algorithm expects single labeled data, whenever multiple optimal actions exist for a state, one of these had to be sampled randomly.
The resulting rule list covers several sectors in the grid where the agent either moves upward or to the right, eventually ending up in the goal.
As standard CN2 cannot take alternative actions into account during the mining process, it cannot distil a simple policy, meaning a policy expressed with few rules, yet having a good performance.

Figure~\ref{fig:grid-phase1} presents the resulting simplified policy of our Set-Valued extension of CN2.
Our algorithm can nicely summarize the essence of the policy into two simple rules, making the agent walk to the right border of the grid first, and afterwards walk upwards to the goal.
Note that the rule mining process did not capture the avoidance of the muddy cells which is an exceptional region in the environment. However, Figure~\ref{fig:grid-phase2} shows that the refinement step that we propose in Section \ref{sec:methods} allows extra rules to be identified, restoring optimal performance of the agent around the muddy region.

Our extension of CN2, able to learn from pairs of states and multiple equally-good labels, is only one component of our explainable reinforcement learning framework. We now fully detail the iterative algorithm that we built on our CN2 extension, that allows a policy to be learned, distilled, evaluated and fine-tuned by the user, to ensure that state-action pairs that are crucial for the performance of the policy are classified correctly.

\section{Explaining Policies through Distillation}\label{sec:methods}
In this work, we are using inductive rule mining to translate a deep reinforcement learning policy into a simplified ordered list of  rules that reveal the agent's behaviour in the environment.
The simplification is performed through a rule-mining approach that exploits the meta-information that is present in the RL learning algorithm, hereby balancing between accuracy and readability.

We extended the implementation of CN2 in the open source data mining framework Orange\footnote{\url{https://github.com/biolab/orange3}}, in order to support multi-label data and incorporated our proposed $\text{WRA}_{set}$ heuristic to perform set-valued rule mining.
We evaluated our methods on an altered level of the Mario AI benchmark~\cite{Karakovskiy2012}.
The goal of the Mario agent is to collect rewarding red coins and avoiding penalising green and blue coins.
The action space of the Mario agent is restricted so that it can only press one of the 5 available controller buttons during each step.
The agent can also decide not to press any controller button, bringing the total number of actions to 6.

\begin{figure}[!htbp]
    \centering
    \begin{tikzpicture}
        \matrix[matrix of nodes, nodes={anchor=center},column sep=-2mm, ampersand replacement=\&, column 1/.style={anchor=base west}]{
            \includegraphics[width=.40\columnwidth]{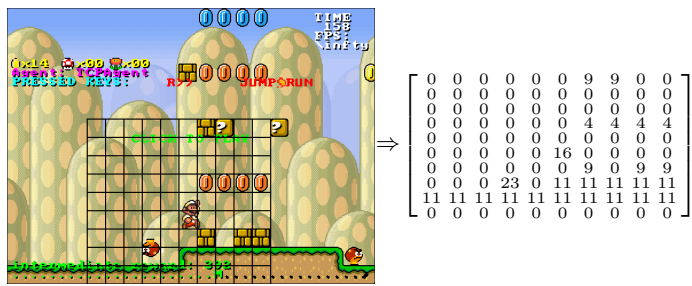}
            \&
            $\Rightarrow$
            \&
            $\begin{bsmallmatrix}
                0 & 0 & 0 & 0 & 0 & 0 & 9 & 9 & 0 & 0\\
                0 & 0 & 0 & 0 & 0 & 0 & 0 & 0 & 0 & 0\\
                0 & 0 & 0 & 0 & 0 & 0 & 0 & 0 & 0 & 0\\
                0 & 0 & 0 & 0 & 0 & 0 & 4 & 4 & 4 & 4\\
                0 & 0 & 0 & 0 & 0 & 0 & 0 & 0 & 0 & 0\\
                0 & 0 & 0 & 0 & 0 & 16 & 0 & 0 & 0 & 0\\
                0 & 0 & 0 & 0 & 0 & 0 & 9 & 0 & 9 & 9\\
                0 & 0 & 0 & 23 & 0 & 11 & 11 & 11 & 11 & 11\\
                11 & 11 & 11 & 11 & 11 & 11 & 11 & 11 & 11 & 11\\
                0 & 0 & 0 & 0 & 0 & 0 & 0 & 0 & 0 & 0\\
            \end{bsmallmatrix}$
            \\
        };
    \end{tikzpicture}
    \caption{Transformation from a MarioAI game frame to a 10 by 10 receptive field grid centred around Mario, which is returned by the emulator. Each cell of the grid corresponds to a pixel region about the size of a brick block in the game. The game objects in each of the grid's cells are translated into numerical labels as state representation for the learning agent. Whenever the receptive field ranges out of the game frame, it is padded with zeros (e.g.~the bottom row).}
    \label{fig:perceptive_field}
\end{figure}

The observations provided to the RL agent are built from a description of the enemies, coins and blocks around the current position of Mario in the environment. 
We consider a 10 by 10 receptive field, that is, we look at 10 by 10 unit squares in the environment, centred around Mario's current position. 
Each square can either be empty, or contain a block, enemy or coin. Because every class of enemy, or every colour of a coin, is uniquely identified, each square can take 24 different values (0 being empty, then positive values identifying the kind of object in the square). 
When producing observations for the agent, we consider this 10 by 10 set of squares that can each take a value among 24, and produced a flattened, one-hot encoded, large observation vector of $10 \times 10 \times 24 = 2400$ entries, with each entry either 0 or 1.
Figure~\ref{fig:perceptive_field} illustrates the transformation from a game frame to a receptive field. Because one-hot encoded states can be decoded back to their integer variant (a 10 by 10 grid of integers in our case), and the integers we consider uniquely identify objects in the Mario game that have human-friendly game, any observation passed to the agent can also be described in a human-friendly way, such as ``\textit{there is a red coin above Mario}''.

As explained above, we propose a two-phased distillation algorithm, to produce meaningful rules from a black-box policy learned with Deep Reinforcement Learning.
The first phase produces a list of rules that approximate how the Deep RL policy maps states to actions.
These rules capture the general and high-level behaviour of the policy, and produce compact and human-friendly descriptions. 
However, some of the approximations made by our rule mining algorithm may lead to rules that, if executed in the environment, lead to a significant loss in performance. Because small mistakes in a single state may or may not have an impact on the performance of the agent, increasing the target accuracy of the rule-mining algorithm does not address the problem. Instead, we introduce a second phase, that evaluates the rules in the environment, and corrects small mistakes when they have an impact on the performance of the agent.

\subsection{Phase 1: Supervised post-processing}\label{subsec:phase1}

Our first phase distills a Deep Reinforcement Learning policy into rules, using the rule-mining algorithm that we introduce in Section \ref{sec:set-valued-rule-mining}:

\begin{enumerate}
    \item We train a Deep RL agent, that we assume produces a stochastic policy. Q-Values can be mapped to a stochastic policy with Boltzmann exploration;
    \item We record multiple trajectories sampled from the policy;
    \item We translate the trajectories into a collection of symbolic states and set-valued actions, so that each state can be mapped to several actions if the Deep RL policy gives similar probability to these actions;
    \item Our rule-mining algorithm learns rules that cover the state-action pairs identified above.
\end{enumerate}

First, a deep RL agent is trained to control Mario to play a single level.
In this paper we realise this through the use of Bootstrapped Dual Policy Iteration (BDPI)~\cite{Steckelmacher2019}. 
We chose this algorithm because it produces an explicit actor, from which we can obtain action probabilities, and the off-policy actor-critic nature of the algorithm leads to an actor that computes the probability that an action is optimal, instead of a simple number that empirically led to high returns, as Policy Gradient approaches do.

Once a policy has been learnt, a data set is recorded by executing the converged RL policy in the environment for 50 episodes.
The data set consists of tuples $(s, \pi(s))$ of state observations and action probabilities for each step of the performed episodes.\footnote{Since BDPI is an actor-critic algorithm, we use the actor predictions as output, i.e.~probabilities over actions. With value-based algorithms, e.g.~DQN~\cite{Mnih2015}, one could record the Q-values instead.}

Given the data set of states and their corresponding action probabilities, a translation is applied in order to prepare for mining the rules.
The numerical values of the state observations are transformed back to their symbolic counterpart, as each value in the observation represents a specific game object in Mario's receptive field.
Since the state observation represents a receptive field around Mario, we can assign x,y coordinates to each cell (feature of the state) with the agent in the center.

For each state observation $s$ in the data set, the predicted action distribution $\pi(s)$ is transformed into a set, $\mathcal{L}(s)$.
This set denotes which actions are considered suitable to be executed in that particular state.
Equation~\ref{eq:action-prob-transform} describes this transformation formally.
For each state $s$ in the recordings, we take the probability of the best action, i.e.~the action with the highest probability, $\max_{a'}(\pi(s,a'))$, and set a proportional threshold, $\tau * \max_{a'}\pi(s,a')$, where $\tau \in (0,1]$ determines how tolerant we are with respect to sub-optimal actions.
Lower values of $\tau$ will lead to more actions exceeding the threshold, while high values will put the threshold closer to $\max_{a'}(\pi(s,a'))$.
For our experiments we picked a value for $\tau = 90\%$.

\begin{equation} \label{eq:action-prob-transform}
    \forall s\in S: \mathcal{L}(s) = \{ a\in A(s)\ : \pi(s,a) \geq \tau * \max_{a'}(\pi(s,a')) ~|~ \forall a' \in A(s) \}
\end{equation}

Once the symbolic data set is created, we can proceed mining rules with our Set-Valued CN2 algorithm.
We explored different settings; providing the rule mining the full state space of the RL as feature space or reducing the feature space to the upper right quadrant of Mario's receptive field, that is the cells in front of Mario. We also performed rule mining including and excluding the inequality ($!=$) predicate. The reduced feature set, with the excluded inequality predicate provides more interpretable rules, and are listed below. 
The default parameters of the beam search, expressing the maximal number of conditions per rule, the minimal number of samples to be covered by a rule and the beam width, were put respectively to 5, 20 and 10.
Learned rules by Set-Valued CN2 using our $\text{WRA}_{set}$ heuristic, that explain the policy our BDPI agent has learned in the Mario environment, are listed below. Note that the rules need to be interpreted sequentially.

\texttt{
\begin{enumerate}
    \item IF (1, 5)=NULL AND (0, 5)=NULL AND (0, 1)=NULL THEN Class=JUMP
    \item IF (1, 5)=BRICK AND (1, 0)=COIN\_RED THEN Class=RIGHT
    \item IF (1, 5)=BRICK AND (2, 4)=COIN\_RED THEN Class=RIGHT 
    \item IF (4, 2)=NULL AND (0, 4)=NULL THEN Class=JUMP
    \item IF (4, 5)=NULL AND (3, 3)=NULL THEN Class=JUMP
    \item IF (3, 5)=BRICK THEN Class=JUMP
    \item IF (2, 2)=BRICK THEN Class=JUMP 
    \item IF (0, 2)=NULL AND (0, 1)=NULL THEN Class=RIGHT
    \item IF (0, 5)=NULL THEN Class=JUMP
    \item IF TRUE THEN Class=JUMP
\end{enumerate}
}

The extraction resulted in a list of 10 rules, capturing the agent's behaviour at a high level.
In overall, the conditions of the rules focus on empty spaces (NULL) and the presence of bricks or red coins. 
The second rule, for instance, explains that the policy likes to move to the right when there is a (positive rewarding) red coin in front of Mario. 
However, a problem with the learned rules is that they over-emphasise jumping, the action that is most often the best one, but that is not enough to finish a Mario level (Mario also has to move right to win). 
While the rules provide a high-level explanation of the policy followed by the agent, they do not allow the agent to obtain high returns when executed. 
Our second phase, the refinement, addresses this issue by allowing the rule mining algorithm to produce more refined rules, that lead to better execution performance.

\subsection{Phase 2: Refinement}\label{subsec:phase2}

The rules produced by the first phase, described above, are optimised for classification accuracy. However, accuracy is not the main objective in policy distillation, as even high accuracy may still lead to poor performance in the environment. We therefore introduce a second phase, that refines the rules of the first phase based on roll-outs in the reinforcement learning environment. The second phase ensures that good policies are produced by our algorithm:

\begin{enumerate}
    \item Use the Deep RL policy to perform an episode in the environment, observe the predicted actions step-wise;
    \item In each time-step, predict the action that would be chosen by the obtained rules from phase 1 and note which rule was applied for the prediction;
    \item When the predicted action of the Deep RL policy and the rules are different, store the observed state together with the action of the Deep RL policy in the training set of the respective rule; 
    \item Once the episode is finished, we extend each training set with samples from phase 1 that were correctly classified in order to get a balanced data set;
    \item Launch a new rule mining process for each balanced training set using Set-Valued CN2 with the respective rule of the first phase as a starting point for the beam search process.
\end{enumerate}

The procedure described above leads to each rule produced during phase 1 to have additional datapoints belonging to them, created when the rules and the black-box policy disagree in some state. By re-running our rule-mining algorithm on these rules, with extra datapoints, overly-general rules that lead to poor performance will be split into additional rules that better capture important actions in important states.
If we were to take for example the fourth rule from the resulting rule list in the first phase, one of the resulting refined rules would be:
\begin{sloppypar}
\texttt{IF (4, 2)=NULL AND (0, 4)=NULL AND (2, 5)!=BRICK AND (0, 3)=NULL AND (1, 2)!=COIN\_RED AND (1, 2)!=COIN\_BLUE AND (1, 2)!=BORDER\_CANNOT\_PASS\_THROUGH THEN Class=RIGHT}
\end{sloppypar}
This additional rule ensures that the agent sometimes moves right, allowing the level to be finished. Because we keep track of which phase-1 rules are split into phase-2 rules, we are able to show the user a hierarchical view of the rules, with simple and general phase-1 rules that capture the overall policy, and some rules having \textit{children} that identify refinements required for top performance.

\begin{figure}[!htbp]
    \centering
    \includegraphics[width=.5\textwidth]{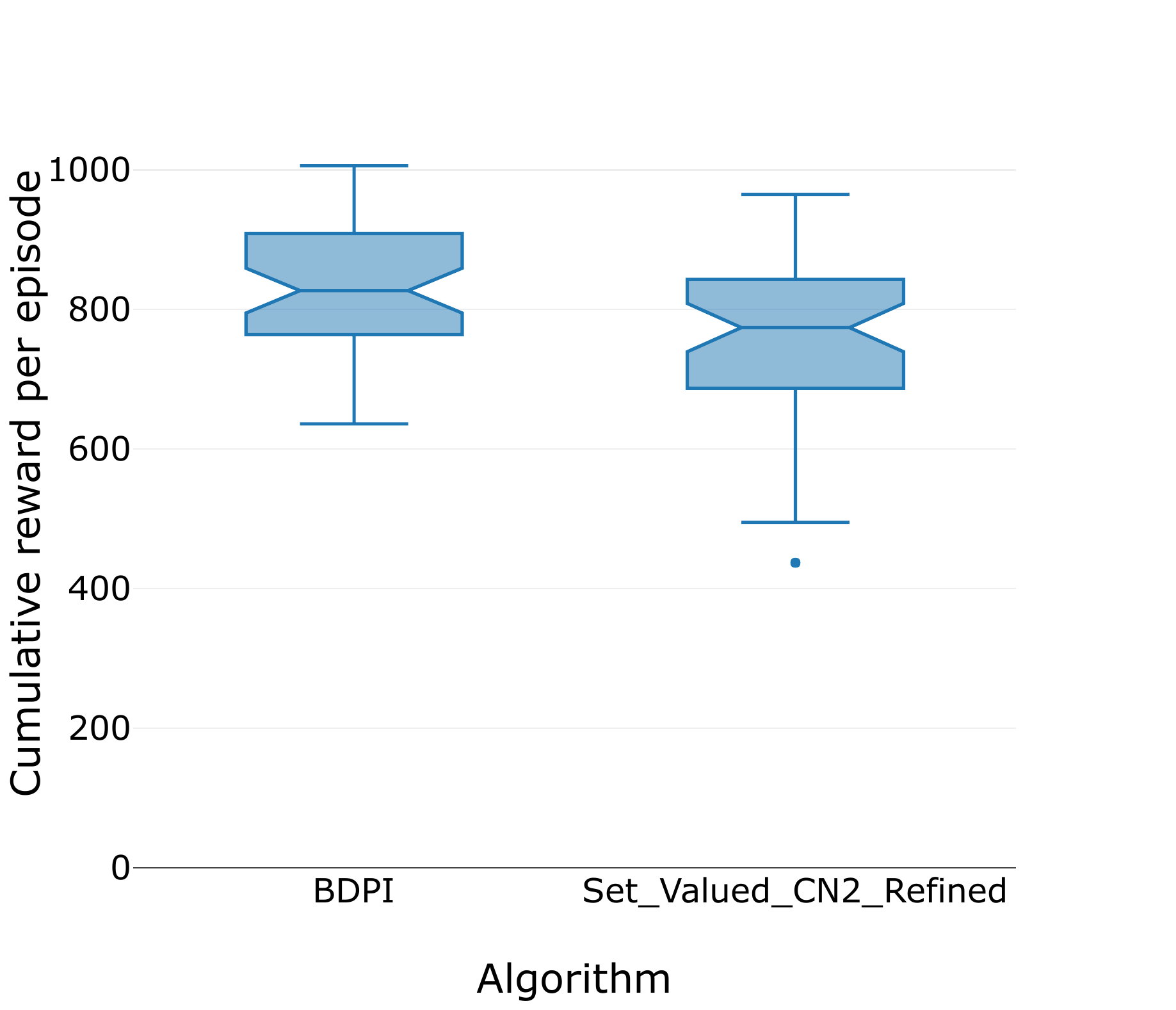}
    \caption{Box plot comparison of episodic cumulative reward (y-axis) achieved by the refined extracted rules of our Set-Valued CN2 on the Mario environment with the original BDPI agent. 50 episodes were played by both agents (x-axis). The refined rules we obtain after phase 2 have an equivalent performance to the original policy since the box plots highly overlap. A bad-performing agent obtains a score 100 on this environment.}
    \label{fig:phase-2-bdpi-vs-set-val-cn2-refined}
\end{figure}

Performing this refinement procedure in our example of Mario with the rules obtained from the first phase (listed in Subsection~\ref{subsec:phase1}), caused a refinement of multiple rules.
We executed the refined rule list in the environment for 50 episodes and compared its performance with the performance of the original BDPI policy. Figure~\ref{fig:phase-2-bdpi-vs-set-val-cn2-refined} shows that the refined rules are capable to reach a performance similar to BDPI, which demonstrates that our algorithms are able to produce intuitive explanations of black-box reinforcement learning policies, while ensuring that the explanations correspond to high-performance policies.

\section{Conclusion}\label{sec:conclusion}
In this paper we have proposed an approach to translate a policy trained by a Deep RL algorithm into a set of rules. 
The set is human-readable, captures the behaviour of the policy, and can be iteratively refined to obtain higher returns if executed in the environment. 
A key element of our approach is that we associate every state to a \emph{set} of equally-good actions, obtained by looking at the action probabilities predicted by the RL policy. 
This part of our contributions allows our modified rule mining algorithm to simplify its rule set.
By allowing to select one of the almost equally-good actions for a state, the number of generated rules is significantly reduced compared to applying CN2 on the greedy policy. 
We find the iterative refinements appealing with respect to explainability. 
The rules of phase one are not altered, but only hierarchically expanded. 
Thus, the rules of the first phase provide a high-level view and every refinement adds details. 
Our refinement process is driven by the performance of the rules, measured by their execution in the environment. 
This prevents us from being dependent on magic numbers (i.e.~the setting of the different parameters) of the rule mining algorithm.  
Furthermore, our method only assumes that the Deep RL algorithm produces a probability distribution over the actions, which can also be retrieved from value-based algorithms by using e.g.~a Softmax function over Q-values. 
Alternatively, Equation~\ref{eq:action-prob-transform} can also be directly applied to Q-values. 
Finally, in our experiments, the states are semantically meaningful for users. The readability of the rules could however be further improved by using rule simplification or \emph{syntactic sugar}. For example, in the Mario experiment, position (0, 4) could be translated into \emph{right of Mario}. In settings where the features are not semantically meaningful, a double representation could be used as proposed in~\cite{de_giacomo_restraining_2020}, where states represented in the feature space of the RL agent are grounded in a semantically meaningful representation for users. In future work we will also explore other rule-mining algorithms, as the use of CN2 is not crucial to our approach. What is however important is that the algorithm can handle Set-Valued labels. Therefore, adaptations of bi-clustering or Markov Random fields are appealing for exploiting meta-information of an RL process.

\subsubsection*{Acknowledgements.}
This work is supported by the Research Foundation Flanders (FWO) [grant numbers G062819N and 1129319N], the AI Research Program from the Flemish Government (Belgium) and the Francqui Foundation. This work is part of the research program Hybrid Intelligence with project number 024.004.022, which is (partly) financed by the Dutch Ministry of Education, Culture and Science (OCW).

\bibliographystyle{splncs04}
\bibliography{references}

\begin{thebibliography}{10}
\providecommand{\url}[1]{\texttt{#1}}
\providecommand{\urlprefix}{URL }
\providecommand{\doi}[1]{https://doi.org/#1}

\bibitem{agogino_analyzing_2008}
Agogino, A.K., Tumer, K.: Analyzing and visualizing multiagent rewards in
  dynamic and stochastic domains. Autonomous Agents and Multi-Agent Systems
  \textbf{17}(2),  320--338 (2008). \doi{10.1007/s10458-008-9046-9}

\bibitem{alharin_reinforcement_2020}
Alharin, A., Doan, T.N., Sartipi, M.: Reinforcement {Learning} {Interpretation}
  {Methods}: {A} {Survey}. IEEE Access  \textbf{8},  171058--171077 (2020).
  \doi{10.1109/ACCESS.2020.3023394}

\bibitem{Brys2014}
Brys, T., Now\'{e}, A., Kudenko, D., Taylor, M.E.: Combining multiple
  correlated reward and shaping signals by measuring confidence. In:
  Proceedings of the Twenty-Eighth AAAI Conference on Artificial Intelligence.
  p. 1687–1693. AAAI Press, Palo Alto, California (2014)

\bibitem{clark_cn2_1989}
Clark, P., Niblett, T.: The {CN2} induction algorithm. Machine Learning
  \textbf{3}(4),  261--283 (1989). \doi{10.1007/BF00116835}

\bibitem{Coppens2019a}
Coppens, Y., Efthymiadis, K., Lenaerts, T., Nowé, A.: Distilling {Deep}
  {Reinforcement} {Learning} {Policies} in {Soft} {Decision} {Trees}. In:
  Miller, T., Weber, R., Magazzeni, D. (eds.) Proceedings of the {IJCAI} 2019
  {Workshop} on {Explainable} {Artificial} {Intelligence}. pp.~1--6. Macau
  (2019)

\bibitem{de_giacomo_restraining_2020}
De~Giacomo, G., Iocchi, L., Favorito, M., Patrizi, F.: Restraining {Bolts} for
  {Reinforcement} {Learning} {Agents}. In: Proceedings of the Thirty-Fourth
  AAAI Conference on Artificial Intelligence. vol.~9, pp. 13659--13662. AAAI
  Press, Palo Alto, California (2020). \doi{10.1609/aaai.v34i09.7114}

\bibitem{Frosst2017}
Frosst, N., Hinton, G.: Distilling a {Neural} {Network} {Into} a {Soft}
  {Decision} {Tree}. In: Besold, T.R., Kutz, O. (eds.) Proceedings of the
  {First} {International} {Workshop} on {Comprehensibility} and {Explanation}
  in {AI} and {ML} 2017. AI*IA Series, vol.~2071. CEUR Workshop Proceedings,
  Aachen, Germany (2017)

\bibitem{furnkranz_foundations_2012}
Fürnkranz, J., Gamberger, D., Lavrač, N.: Foundations of {Rule} {Learning}.
  Cognitive {Technologies}, Springer, Berlin, Heidelberg (2012).
  \doi{10.1007/978-3-540-75197-7}

\bibitem{gevaert_distillation_2019}
Gevaert, A., Peck, J., Saeys, Y.: Distillation of {Deep} {Reinforcement}
  {Learning} {Models} using {Fuzzy} {Inference} {Systems}. In: Beuls, K.,
  Bogaerts, B., Bontempi, G., Geurts, P., Harley, N., Lebichot, B., Lenaerts,
  T., Louppe, G., Van~Eecke, P. (eds.) Proceedings of the 31st {Benelux}
  {Conference} on {Artificial} {Intelligence} ({BNAIC} 2019) and the 28th
  {Belgian} {Dutch} {Conference} on {Machine} {Learning} ({Benelearn} 2019).
  vol.~2491. CEUR Workshop Proceedings, Aachen, Germany (2019)

\bibitem{Hinton2015distill}
Hinton, G., Vinyals, O., Dean, J.: Distilling the {Knowledge} in a {Neural}
  {Network}. arXiv e-prints arXiv:1503.02531 (2015)

\bibitem{huang_interpretable_2020}
Huang, J., Angelov, P.P., Yin, C.: Interpretable policies for reinforcement
  learning by empirical fuzzy sets. Engineering Applications of Artificial
  Intelligence  \textbf{91} (2020). \doi{10.1016/j.engappai.2020.103559}

\bibitem{Karakovskiy2012}
Karakovskiy, S., Togelius, J.: The {Mario} {AI} {Benchmark} and {Competitions}.
  IEEE Transactions on Computational Intelligence and AI in Games
  \textbf{4}(1),  55--67 (2012). \doi{10.1109/TCIAIG.2012.2188528}

\bibitem{lavrac_rule_1999}
Lavra{\v{c}}, N., Flach, P., Zupan, B.: Rule {Evaluation} {Measures}: {A}
  {Unifying} {View}. In: D{\v{z}}eroski, S., Flach, P. (eds.) Inductive {Logic}
  {Programming}. Lecture {Notes} in {Computer} {Science}, vol.~1634, pp.
  174--185. Springer, Berlin, Heidelberg (1999).
  \doi{10.1007/3-540-48751-4\_17}

\bibitem{Libin2020SEIR}
Libin, P., Moonens, A., Verstraeten, T., Perez~Sanjines, F.R., Hens, N., Lemey,
  P., Nowé, A.: Deep reinforcement learning for large-scale epidemic control.
  In: Proceedings of the Adaptive and Learning Agents Workshop 2020 (ALA2020)
  at AAMAS (2020)

\bibitem{madumal_explainable_2020}
Madumal, P., Miller, T., Sonenberg, L., Vetere, F.: Explainable reinforcement
  learning through a causal lens. In: Proceedings of the Thirty-Fourth AAAI
  Conference on Artificial Intelligence. vol.~3, pp. 2493--2500. AAAI Press,
  Palo Alto, California (2020). \doi{10.1609/aaai.v34i03.5631}

\bibitem{maes_computational_1987}
Maes, P.: Computational {Reflection}. In: Morik, K. (ed.) GWAI-87 11th German
  Workshop on Artifical Intelligence. Informatik-Fachberichte, vol.~152, pp.
  251--265. Springer, Berlin, Heidelberg (1987).
  \doi{10.1007/978-3-642-73005-4\_27}

\bibitem{miller_explanation_2019}
Miller, T.: Explanation in artificial intelligence: {Insights} from the social
  sciences. Artificial Intelligence  \textbf{267},  1--38 (2019).
  \doi{10.1016/j.artint.2018.07.007}

\bibitem{Mnih2015}
Mnih, V., Kavukcuoglu, K., Silver, D., Rusu, A.A., Veness, J., Bellemare, M.G.,
  Graves, A., Riedmiller, M., Fidjeland, A.K., Ostrovski, G., Petersen, S.,
  Beattie, C., Sadik, A., Antonoglou, I., King, H., Kumaran, D., Wierstra, D.,
  Legg, S., Hassabis, D.: Human-level control through deep reinforcement
  learning. Nature  \textbf{518}(7540),  529--533 (2015).
  \doi{10.1038/nature14236}

\bibitem{molnar_interpretable_2019}
Molnar, C.: Interpretable Machine Learning. Leanpub, Victoria, Canada (2019)

\bibitem{Rusu2016}
Rusu, A.A., Colmenarejo, S.G., Gulcehre, C., Desjardins, G., Kirkpatrick, J.,
  Pascanu, R., Mnih, V., Kavukcuoglu, K., Hadsell, R.: Policy {Distillation}.
  In: {International} {Conference} on {Learning} {Representations} (2016),
  arXiv:1511.06295

\bibitem{rucksties_exploring_2010}
Rückstieß, T., Sehnke, F., Schaul, T., Wierstra, D., Sun, Y., Schmidhuber,
  J.: Exploring parameter space in reinforcement learning. Paladyn, Journal of
  Behavioral Robotics  \textbf{1}(1),  14--24 (2010).
  \doi{10.2478/s13230-010-0002-4}

\bibitem{Steckelmacher2019}
Steckelmacher, D., Plisnier, H., Roijers, D.M., Now{\'e}, A.: Sample-efficient
  model-free reinforcement learning with off-policy critics. In: Brefeld, U.,
  Fromont, E., Hotho, A., Knobbe, A., Maathuis, M., Robardet, C. (eds.) Machine
  Learning and Knowledge Discovery in Databases. Lecture Notes in Computer
  Science, vol. 11908, pp. 19--34. Springer, Cham, Switzerland (2020).
  \doi{10.1007/978-3-030-46133-1\_2}

\bibitem{Sutton2018}
Sutton, R.S., Barto, A.G.: Reinforcement {Learning} : {An} {Introduction}. MIT
  Press, Cambridge, Massachusetts, 2nd edn. (2018)

\bibitem{Sutton2000}
Sutton, R.S., McAllester, D., Singh, S., Mansour, Y.: {Policy Gradient Methods
  for Reinforcement Learning with Function Approximation}. In: Neural
  Information Processing Systems ({NIPS}). pp. 1057--1063 (2000)

\bibitem{tadepalli_relational_2004}
Tadepalli, P., Givan, R., Driessens, K.: Relational {Reinforcement} {Learning}:
  {An} {Overview}. In: Tadepalli, P., Givan, R., Driessens, K. (eds.)
  Proceedings of the {ICML}-2004 {Workshop} on {Relational} {Reinforcement}
  {Learning}. pp.~1--9. Banff, Canada (2004)

\bibitem{todorovski_predictive_2000}
Todorovski, L., Flach, P., Lavrač, N.: Predictive {Performance} of {Weighted}
  {Relative} {Accuracy}. In: Zighed, D.A., Komorowski, J., Żytkow, J. (eds.)
  Principles of {Data} {Mining} and {Knowledge} {Discovery}. Lecture {Notes} in
  {Computer} {Science}, vol.~1910, pp. 255--264. Springer, Berlin, Heidelberg
  (2000). \doi{10.1007/3-540-45372-5\_25}

\bibitem{zambaldi2018deep}
Zambaldi, V., Raposo, D., Santoro, A., Bapst, V., Li, Y., Babuschkin, I.,
  Tuyls, K., Reichert, D., Lillicrap, T., Lockhart, E., Shanahan, M., Langston,
  V., Pascanu, R., Botvinick, M., Vinyals, O., Battaglia, P.: Deep
  reinforcement learning with relational inductive biases. In: International
  Conference on Learning Representations (2019)

\end{thebibliography}
\end{document}